\documentclass[nohyperref]{article}
\usepackage{PRIMEarxiv}

\usepackage[utf8]{inputenc} % allow utf-8 input
\usepackage[T1]{fontenc}    % use 8-bit T1 fonts
\usepackage{hyperref}       % hyperlinks
\usepackage{url}            % simple URL typesetting
\usepackage{booktabs}       % professional-quality tables
\usepackage{amsfonts}       % blackboard math symbols
\usepackage{nicefrac}       % compact symbols for 1/2, etc.
\usepackage{microtype}      % microtypography
\usepackage{lipsum}
\usepackage{fancyhdr}       % header
\usepackage{graphicx}       % graphics
\graphicspath{{media/}}     % organize your images and other figures under media/ folder

\usepackage{microtype}
\usepackage{graphicx}
\usepackage{subfigure}
\usepackage{booktabs} % for professional tables

\usepackage{hyperref}

\usepackage{amsmath}
\usepackage{amssymb}
\usepackage{mathtools}
\usepackage{amsthm}
\usepackage{listings}
\usepackage[detect-all]{siunitx}
\usepackage[capitalize,noabbrev]{cleveref}

\newcommand{\etal}{et~al. }

\theoremstyle{plain}

\theoremstyle{definition}

\theoremstyle{remark}

\usepackage[textsize=tiny]{todonotes}

\title{General Cyclical Training of Neural Networks
}

\author{
	Leslie N. Smith \\
	U.S. Naval Research Laboratory, \\
	Naval Center for Applied Research in AI, Code 5514 \\
	4555 Overlook Ave., SW., Washington, D.C.  20375 \\
	\texttt{leslie.smith@nrl.navy.mil} \\
}

\begin{document}
	\maketitle

\begin{abstract}
This paper describes the principle of ``General Cyclical Training'' in machine learning, where training starts and ends with ``easy training'' and the ``hard training'' happens during the middle epochs. We propose several manifestations for training neural networks, including algorithmic examples (via hyper-parameters and loss functions), data-based examples, and model-based examples. Specifically, we introduce several novel techniques: cyclical weight decay, cyclical batch size, cyclical focal loss, cyclical softmax temperature, cyclical data augmentation, cyclical gradient clipping, and cyclical semi-supervised learning.  In addition, we demonstrate that cyclical weight decay, cyclical softmax temperature, and cyclical gradient clipping (as three examples of this principle) are beneficial in the test accuracy performance of a trained model.   Furthermore, we discuss model-based examples (such as pretraining and knowledge distillation) from the perspective of general cyclical training and recommend some changes to the typical training methodology. In summary, this paper defines the general cyclical training concept and discusses several specific ways in which this concept can be applied to training neural networks.
In the spirit of reproducibility, the code used in our experiments is available at \url{https://github.com/lnsmith54/CFL}.
\end{abstract}

\section{Introduction}
\label{sec:intro}

Deep neural networks lie at the heart of many of the artificial intelligence applications that are ubiquitous in our society.   Over the past several years, methods for training these networks have become more automatic \cite{awad2021dehb,eggensperger2021hpobench,morales2021survey,bischl2021hyperparameter,dong2021automated} but still remain more an art than a science.  This paper introduces the high-level concept of general cyclical training as another step in making it easier to optimally train neural networks.  We argue that many of the settings that are held constant throughout training need not be and training improves when they are not constant.

We define general cyclical training as any collection of settings where the training starts and ends with ``easy training'' and the ``hard training'' happens during the middle epochs.
In other words, it can be considered as a combination of curriculum learning \cite{bengio2009curriculum} in the early epochs with fine-tuning toward the end of training, plus training over the full problem space for greater generalization happening during the middle epochs.  
General cyclical training is analogous to curriculum learning in the sense that numerous specific techniques can embody its principles.

It has been shown that many important aspects of neural network learning take place within the very earliest iterations or epochs of training  \cite{golatkar2019time,frankle2020early}.  It is best to construct neural network training such that the network’s weight updates during the earliest epochs are relatively easy and of the highest quality for the task. This first part of a network’s training could use a curriculum learning approach.  As the training proceeds, one increases the learning to span the full problem space and the hard work of learning to generalize is achieved during the middle epochs.   The final epochs of the training should fine-tune the model on the desired data or tasks, because this is when the network learns the more complex patterns \cite{you2019does} from the most relevant training samples.  

Based on the above intuition, this paper proposes that cyclical approaches for training can be generalized to all aspects of neural network training.  In addition to learning rates, cyclical training can extend to other hyper-parameters, such as weight decay and batch size.  In addition to hyper-parameters, this approach can be extended to loss functions, data-based methods, and model-based methods.
For example, data augmentation methods can be cyclical by using no augmentation or weak augmentations early in the training cycle, then adding complex augmentations as the training proceeds, and eliminating augmentations in the later part of training.  In addition, general cyclical training answers the question, ``Which samples should be learned first?'' In most scenarios, one should learn the easy samples first and the hard ones during the middle epochs.  
%However, this general rule does not apply in the case of significant class imbalance.

Adaptive hyper-parameters during training have become common.  Cyclical learning rates \cite{smith2017cyclical}, one cycle learning rates \cite{smith2019super}, and cosine annealing with warm restarts \cite{loshchilov2016sgdr} have been accepted by the deep learning community and incorporated in PyTorch.  General cyclical training provides an intuitive understanding for the value of a one cycle training regime.   Furthermore, this idea of allowing a hyper-parameter value to change during training has been extended to other hyper-parameters, such as weight decay \cite{zhang2018three,bjorck2020understanding,nakamura2019adaptive,lewkowycz2020training,smith2019diet} and batch sizes \cite{smith2017don}.

In summary, the concept of general cyclical training is to start training in a simpler fashion during the early epochs, to train within the entire problem space and challenging conditions during the middle epochs, and to finish with fine-tuning on the most confident samples.  General cyclical training includes adapting any and all factors that impact the network's training (i.e., hyper-parameters, data, loss functions).  
%It can follow any reasonable schedule, even though we limit ourselves to a linear schedule in this paper due to simplicity.

\section{General Cyclical Training}
\label{sec:GCT}

When training in machine learning, and especially with neural networks, there are several settings that the practitioners must make that impact the final performance of the model.  There are decisions regarding the hyper-parameters and the training samples that impact not only the ease of the training, but also the final generalization performance.  
%For clarity, we limit our comments in this paper to the training of neural networks but the concepts do apply to the training of other machine learning methods.

More formally, let us define the set of training settings $ P = \{ (P_j) : j \in (0,1, . . . , n)  \},$ where $ P_0 $ can be a hyper-parameter value, a subset of the data, or a set of any combination that is easiest for the network to learn, while $ P_n $ represents the set of training environmental conditions that is difficult for the network to learn. 
%These difficult training settings improve the generalization performance of the trained network.  
In practice, one chooses a single $P_j$ as a trade-off that provides the best performance.  A cyclical approach improves on this trade-off by using a range of settings: at the beginning of the training, use a smaller $j$ to jump-start the learning, shift gradually to $P_j$ for a larger $j$ during the middle of the training, and followed that with settings for which there is a decrease of $j$ to the end of training.

%-------------------------------------------------------------------------
\begin{figure}[htb]
	%	\vskip 0.2in
	\begin{center}
		\centerline{\includegraphics[width=0.6\columnwidth]{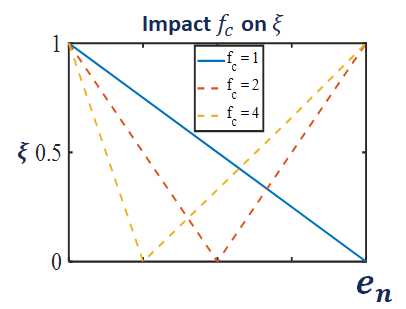}}
		\caption{\textbf{Effect of $f_c$ on the cyclical schedule:} This figure shows the impact of different values for $f_c$ on the cyclical schedule. 
		}
		\label{fig:f_c}
	\end{center}
	\vskip -0.2in
\end{figure}
%-------------------------------------------------------------------------

The general structure of cyclical training can proceed with any schedule over some range of $j$ for $ P = \{ (P_j) : j \in (0,1, . . . , n)  \}$, but for the sake of simplicity, we limit our comments to a linear schedule from $P_i$ to $P_m$ according to
\begin{equation}
	\xi P_i + (1 - \xi) P_{m}
	\label{eqn:GCT}
\end{equation}
where $\xi$ is defined over the training for a number of epochs $e_n$ as:
\begin{equation}
	\xi = \begin{cases}
		1 - f_c \frac{e_k}{e_n}      & \quad \text{if } f_c \times e_k \leq e_n \\
%		0      & \quad \text{else if } f_c = 0 \\
		\big( f_c \frac{e_k}{e_n} - 1\big)  / (f_c - 1)  & \quad \text{ otherwise}
	\end{cases}
	\label{eqn:xi}
\end{equation}
where $e_k$ corresponds to the current training epoch number.
Here, we introduce a cyclical factor $f_c$ that generalizes the shape of the cycle.  If $f_c = 1$, Equation \ref{eqn:GCT} goes from the $P_i$ at the beginning of the training to $P_m$ at the end  (see Figure \ref{fig:f_c}).  If $f_c = 2$, the shape for $\xi$ resembles an upside down equilateral triangle (i.e., going from $P_i$ to $P_m$ in the first half of training and from $P_m$ to $P_i$ in the second half).  If $f_c = 4, P_m$,  is reached at a quarter of the way through the training and then linearly decreases from $P_m$ to  $P_i$ for the remaining epochs.
Although we don't discuss any other schedule, one can imitate the more complex learning rate schedules, such as stochastic gradient descent with warm restarts (SGDR) \cite{loshchilov2016sgdr} or a polynomial schedule.

Furthermore, we use $P_i$ instead of $P_0$ and  $P_m$ instead of $P_n$ where  $i \geq 0$ and  $m \leq n$ to incorporate the flexibility not to include in the training regime settings that are too easy or too difficult.

\section{Algorithmic Examples}
\label{sec:algo}

Algorithmic examples primarily include the use of dynamic hyper-parameters and various loss functions, which are discussed in this Section.

\subsection{Hyper-parameters and Regularization}
\label{subsec:HP}

The use of a decaying learning rate schedule (where the learning rate value is reduced during the training) is standard practice for network training but the use of adaptive hyper-parameters during training has become common. 
For example, the use of cyclical learning rates \cite{smith2017cyclical,smith2019super,loshchilov2016sgdr} has become widely accepted by the deep learning community.  Unlike the decaying learning rate schedule, cyclical learning rates start with a small value of the learning rate, which enables the network's weights to move toward a good direction in the loss landscape \cite{li2017visualizing} and increases the learning rate in the early epochs.
In addition, learning rate warmup \cite{goyal2017accurate} is essentially equivalent to cyclical learning rates, although the warm up period is often restricted to a few of the early epochs (i.e., the same as setting $f_c$ in Equation \ref{eqn:xi} to a large value).  

This idea of allowing a hyper-parameter value to change during training (in replacement for a learning rate schedule) has been extended to other hyper-parameters, such as weight decay \cite{zhang2018three,bjorck2020understanding,nakamura2019adaptive,lewkowycz2020training} and batch sizes \cite{smith2017don}.
Previous work \cite{smith2019diet,smith2017don} has demonstrated empirically a relationship between the optimal hyper-parameters of learning rate (LR), weight decay (WD), batch size (BS), and momentum (m) as
\begin{equation}
	\frac{LR \times WD}{BS \times (1 - m)} \simeq 10^{-6}.
	\label{eqn:HPratio}
\end{equation}

Equation \ref{eqn:HPratio} and the success of cyclical learning rates implies that a cyclical approach also might work for weight decay, batch size, and momentum as well.
That is, it might be easier for the network to learn when weight decay or momentum starts smaller or if batch size starts larger, followed by a larger weight decay/momentum or smaller batch size in the middle epochs.  While \cite{smith2017don} proposed increasing batch size instead of decreasing the learning rate, we propose taking this a step further to a cyclical batch size. 

%====================================================================================
\begin{table*}[bt]
	\caption{\textbf{Cyclical Weight Decay: } Top-1 test classification accuracies comparing cyclical weight decay (CWD) to constant weight decay for CIFAR-10, 4K CIFAR-10 (i.e., only 4000 training samples),  CIFAR-100, and ImageNet. In all of these experiments, CWD improved on the network's performance as compared to training with a constant weight decay.}
	\label{tab:WD}
	\vskip 0.15in
	\begin{center}
		\begin{small}
			\begin{sc}
				\begin{tabular}{|l|c|c|c|c|}
					%					\toprule
					\hline
					Data set  & Accuracy  & Accuracy  & Accuracy  & Accuracy \\
					\hline
					%					Model &  &  &  &   &   \\
					%					\midrule
					CIFAR-10  & 97.33$\pm$ 0.07 & 97.36$\pm$ 0.02  &  97.45$\pm$ 0.11&  \textbf{97.47$\pm$ 0.06} \\
					WD range  & $ 5 \times 10^{-4} $  & $4 \times 10^{-4} - 6 \times 10^{-4}$  & $ 2 \times 10^{-4} - 8 \times 10^{-4} $  & $ 10^{-4} - 10^{-3} $   \\
					\hline
					4K CIFAR-10  & 86.68$\pm$ 0.34 & 86.95$\pm$ 0.21  &  87.35$\pm$ 0.03&  \textbf{87.55$\pm$ 0.23} \\
					WD range  & $ 5 \times 10^{-4} $  & $4 \times 10^{-4} - 6 \times 10^{-4}$  & $ 2 \times 10^{-4} - 8 \times 10^{-4} $  & $ 10^{-4} - 10^{-3} $   \\
					\hline
					CIFAR-100  & 83.82$\pm$ 0.26 & 84.27$\pm$ 0.31&  84.44$\pm$ 0.19 & \textbf{84.36$\pm$ 0.12} \\
					WD range  & $ 2 \times 10^{-4} $  & $ 10^{-4} - 3 \times 10^{-4} $ & $ 5 \times 10^{-5} - 5 \times 10^{-4} $  & $ 5 \times 10^{-5} - 7 \times 10^{-4} $   \\
					\hline
					ImageNet  & 80.27$\pm$ 0.01& \textbf{80.50$\pm$ 0.05}  &  80.49$\pm$ 0.01 & 80.41$\pm$ 0.11 \\
					WD range  & $ 2 \times 10^{-5} $  & $ 10^{-5} - 3 \times10^{-5} $ & $ 5 \times 10^{-6} - 5 \times 10^{-5} $  & $ 5 \times 10^{-6} - 7 \times 10^{-5} $   \\
					\hline
					%					\bottomrule
				\end{tabular}
			\end{sc}
		\end{small}
	\end{center}
	\vskip -0.1in
\end{table*}
%====================================================================================

In addition, one can combine small cyclical changes in all four hyper-parameters (i.e., LR, WD, BS, and momentum) because it reduces the amount of hyper-parameter tuning required --- so long as the optimal values of the hyper-parameters are within the range of the cyclical values, good generalization results can be obtained. That is, we found that being close counts in training networks.
Finally, we mention that in our experiments, we found that much of the benefits of cyclical training may be achieved with even one cyclical hyper-parameter.  

As an example of general cyclical training with hyper-parameters and for regularization, we here propose and test cyclical weight decay (CWD).  In CWD, the value for weight decay varies over the course of the training by:
\begin{equation}
	WD = \xi WD_{min} + (1 - \xi) WD_{max},
	\label{eqn:CWD}
\end{equation}
where $\xi$ follows Equation \ref{eqn:xi} and $WD_{min}$ and $WD_{max}$ are user-defined hyper-parameters that specify the range for weight decay.

Table \ref*{tab:WD} compares the test accuracies for cyclical weight decay (CWD) to training with tuned hyper-parameters (with a constant weight decay) and learning rate warmstart and cosine annealing \cite{loshchilov2016sgdr}.
For each dataset in this Table there are two rows: the first row presents the mean test accuracy and the standard deviation over four runs (for ImageNet, this is the mean and standard deviation over two runs), and the second row provides the range of weight decay used in the training.  The second column in the Table provides the results of training with a constant weight decay, and the subsequent columns, show the results of training with an increasing range for weight decay.
In our experiments, we found that the performance was relatively insensitive to the value of $f_c$.

%-------------------------------------------------------------------------
\begin{figure}[htb]
	%	\vskip 0.2in
	\begin{center}
		\centerline{\includegraphics[width=0.8\columnwidth]{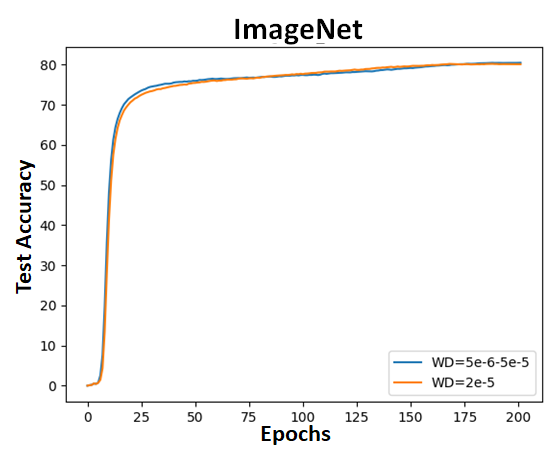}}
		\caption{\textbf{CWD on ImageNet:} This figure shows the ImageNet test accuracy curve during training for cyclical weight decay versus a constant weight decay. The difference between the curves is minor, but shows cyclical weight decay allows a slightly faster rise in test accuracy. 
		}
		\label{fig:imagenet_training}
	\end{center}
	\vskip -0.2in
\end{figure}
%-------------------------------------------------------------------------

The results in Table \ref{tab:WD} show that there is a benefit to training over a range of weight decay values.
For CIFAR-10, using cyclical weight decay improves the network performance relative to using a constant value of $ 5 \times 10^{-3}$, and  the range from $ 10^{-4}$ to $10^{-3} $ has the best performance but using the range from $ 2 \times 10^{-4}$ to $8 \times 10^{-3} $ is within the precision of our experiments.  
The second row of  Table \ref{tab:WD} shows the results when training on only a fraction of the CIFAR-10 training set.  Here we used the first 4,000 samples in the CIFAR-10 training dataset.  Using cyclical weight decay improves the network performance relative to using a constant value of $ 5 \times 10^{-3}$, and  the range from $ 10^{-4}$ to $10^{-3} $ has the best performance.  It is noteworthy that CWD provides a more substantial benefit when the amount of training data is limited.
In addition, the third row of  Table \ref{tab:WD} shows results for CIFAR-100 and the range from $  10^{-4}$ to $8 \times 10^{-4} $  has the best performance.  

%-------------------------------------------------------------------------
\begin{figure}[htb]
	%	\vskip 0.2in
	\begin{center}
		\centerline{\includegraphics[width=0.6\columnwidth]{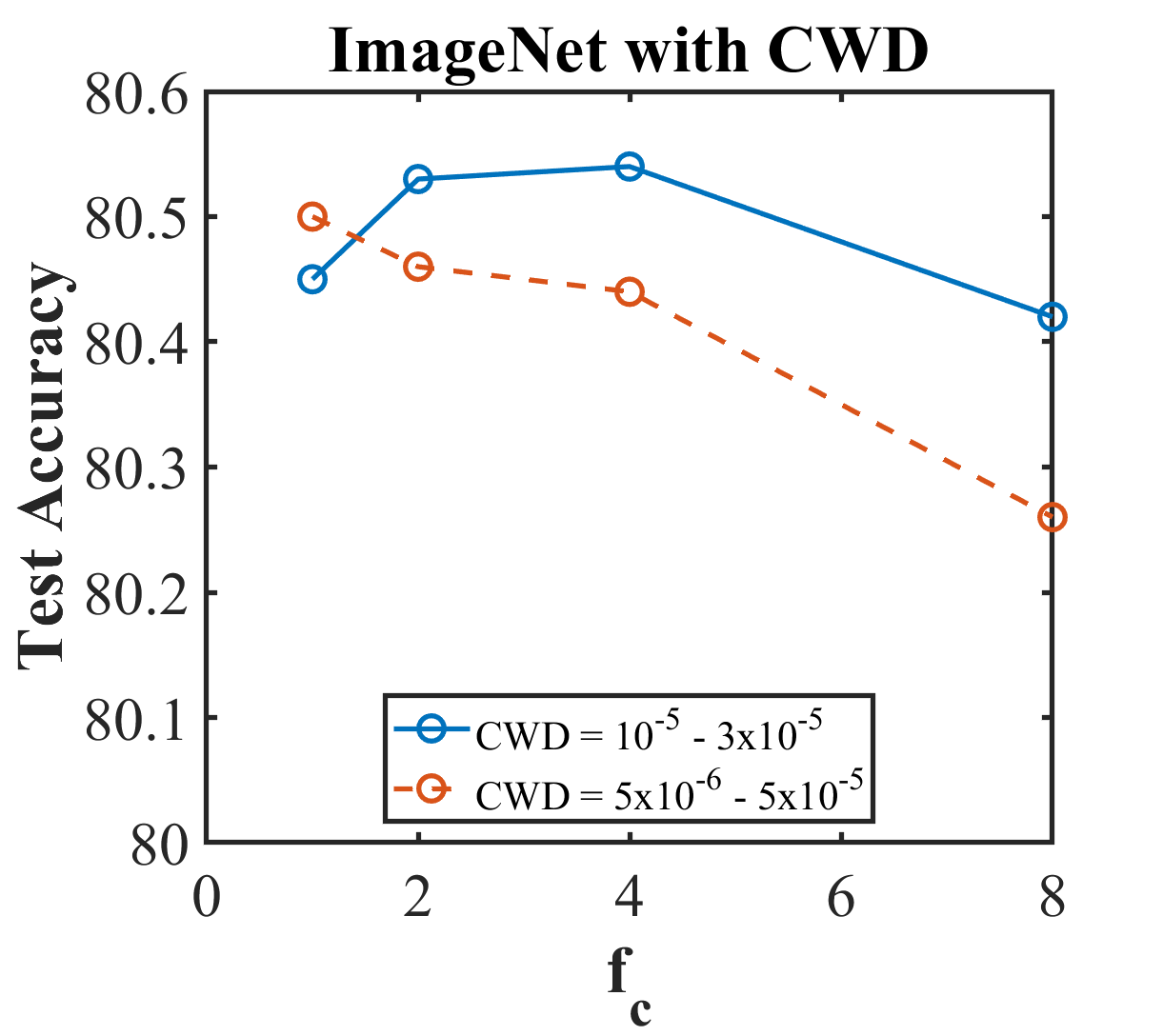}}
		\caption{\textbf{Sensitivity of $f_c$:} This figure shows the ImageNet test accuracy for a range of values for $f_c$ when trained with cyclical weight decay. The accuracy changes are small over this range of $f_c$ values and $f_c = 2$ or 4 are generally good values.
		}
		\label{fig:imagenetcwd}
	\end{center}
	\vskip -0.2in
\end{figure}
%-------------------------------------------------------------------------

%This implies that using a range that includes the optimal weight decay value (in this case $5 \times 10^{-4}$) is best but too much of a range can be harmful.  For this reason, having a range such as $ 10^{-5} - 10^{-2} $ is not a good idea but it also isn't necessary.  Depending on the data, an order of magnitude range is likely to contain the optimal weight decay value.

The fourth row of Table \ref{tab:WD} shows the results of our experiments with CWD on ImageNet.  For ImageNet, the optimal weight decay is $2 \times 10^{-5}$.
Using cyclical weight decay improves the network performance relative to using a constant value and the gain appears to be stable over the small ranges we used in our experiments.
Figure \ref{fig:imagenet_training} compares the test accuracy curves during training for on ImageNet with a constant weight decay versus a range (test accuracy from single imagenet runs are plotted).  While the two curves are similar, the test accuracy rises slightly faster for the cyclical weight decay experiment, which illustrates that the smaller values for weight decay in the earlier epochs allow for slightly faster training than is provided by a constant weight decay.
In addition, note from Table \ref{tab:WD} that the final test accuracy for training with cyclical weight decay is higher than for using a constant weight decay.
 
Cyclical weight decay introduces a new hyper-parameter, $f_c$, in addition to the user-defined $WD_{min}$ and  $WD_{max}$.  Figure \ref{fig:imagenetcwd} shows the test accuracies from training with a range of values for $f_c$ in CWD on ImageNet with a TResNet\_m architecture \cite{ridnik2021tresnet}.  The Figure compares the accuracies when using two different weight decay ranges and the shapes of these curves imply different optimal values for $f_c$.  Actually, these results are mostly within the precision of our experiments, implying that the performance is relatively insensitive to the value of $f_c$.  In Table \ref{tab:WD}, we used a value of $f_c = 4$ for training with CWD.

Implementation of cyclical weight decay is straightforward and is described in the appendix.  Furthermore, PyTorch code is provided at \url{https://github.com/lnsmith54/CFL} to aid in the reproducibility of our experiments.

\subsection{Loss Functions}
\label{subsec:loss}

%In supervised learning with large labeled datasets, even the first epoch training uses the loss from every training sample in updating the weights.  We propose to use only the loss from the easy training samples during the early epochs.  This can be done by a dynamic threshold on the loss, where the loss for samples that exceed the threshold is zeroed or clipped.  Also, the dynamic threshold increases during the first half/part of the training, until all of the training samples are used.  The dynamic threshold decreases in the second part of the training to fine-tune on easier training samples.  Loss clipping is a well-known technique – our novel contribution is to propose a cyclical clipping value during training.

In this Section we describe how the cyclical approaches can be applied to loss functions.
A recent paper proposes a novel cyclical focal loss (CFL) \cite{smith2022cyclical}, which emphasizes the confident samples early and late but focuses on the misclassified examples in the middle of training.  
This is accomplished by introducing a new loss term to the focal loss \cite{lin2017focal} term that more heavily weights confident samples than cross-entropy softmax and is used in the beginning and the end of the training.  In addition, the focal loss term is used during the middle epochs because it more heavily weights hard, less confident samples.
The result is a universal loss function that is superior to cross-entropy softmax and focal loss across balanced, imbalanced, or long-tailed datasets.

%General cyclical training might also be applicable to clipping (start training with a small value that increase the first part of training and decreases late in the training) and semi-supervised learning (start training with only labeled data and increasingly include the unsupervised loss early in training and decrease it late in training).

Another example of a cyclical loss function is possible by making the softmax temperature  \cite{hinton2015distilling} dynamic --- letting the temperature vary over the course of the training.  We call this method cyclical softmax temperature (CST).

Softmax with temperature \cite{hinton2015distilling} can be expressed as:
\begin{equation}
	Softmax(x_i) = \frac{exp(z_i / T)}{\sum_{j=1}^{N} exp(z_j / T) }
	\label{eqn:softmax}
\end{equation}
where $z_i$ is the model's predictions or logits and $T$ is the softmax temperature. Typically, the temperature is 1 in softmax. If the temperature is less than 1, the softmax predictions become more confident.  This is also referred to as ``hard'' softmax probabilities.  If T is greater than 1, then the predictions are less confident (also called ``soft'' probabilities).  
Softer probabilities provide more information from the network as to which classes seem similar to the target class.

%====================================================================================
\begin{table*}[tb]
	\caption{\textbf{Cyclical Softmax Temperature: } CIFAR-10, CIFAR-100 and ImageNet top-1 test  classification accuracies comparing cyclical softmax temperature (CST) to softmax with a temperature = 1.  For all three datasets, CST with a temperature from 0.5 to 2 improved on the network's performance as compared to training with T = 1.  $f_c = 1$ were used in these experiments. }
	\label{tab:CST}
	\vskip 0.15in
	\begin{center}
		\begin{small}
			\begin{sc}
				\begin{tabular}{|l|c|c|c|c|}
					%					\toprule
					\hline
					Data set  & $ T = 1 $ & $ T = 0.75 - 1.5 $  & $ T = 0.5 - 2.0 $ & $ T = 0.33 - 3.0  $   \\
					\hline
					%					Model &  &  &  &   &   \\
					%					\midrule
					CIFAR-10  & 97.33$\pm$ 0.07 &  97.32$\pm$ 0.07 & \textbf{97.43$\pm$0.06}  & 97.28$\pm$ 0.12 \\
					\hline
					4K CIFAR-10  & 86.68$\pm$ 0.34 &  87.14$\pm$0.21 & 87.09$\pm$0.35  & \textbf{87.16$\pm$ 0.54} \\
					\hline
					CIFAR-100  & 83.82$\pm$ 0.26 &  83.91$\pm$ 0.13 & \textbf{83.98$\pm$ 0.25}  & 83.19$\pm$ 0.21 \\
					\hline
					ImageNet  & 80.27$\pm$0.01 & 80.42$\pm$0.12 & \textbf{80.50$\pm$0.04}   &  80.12$\pm$0.13  \\
					\hline
					%					\bottomrule
				\end{tabular}
			\end{sc}
		\end{small}
	\end{center}
	\vskip -0.1in
\end{table*}
%====================================================================================

The intuition for cyclical softmax temperature is that in the first epochs, more confident predictions will help start moving the network's weights in the right direction.  However, in the middle and later epochs, softer probabilities are more appropriate for the remaining training samples.  In addition, softer probabilities reduce the confidence miscalibration of the predictions \cite{guo2017calibration} so that the network's training is based on losses that better match reality.  
%Using a cyclical softmax temperature during training provides an overall more correctly trained network without sacrificing the accuracy gained by training with more confident predictions.

Specifically, we propose a cyclical softmax temperature (CST) where the value for the temperature varies over the course of the training by replacing $T$ in Equation \ref{eqn:softmax} with $T_c$, which we define as:
\begin{equation}
 	T_c = \xi T_{min} + (1 - \xi) T_{max}
	\label{eqn:CWD}
\end{equation}
where $\xi$ follows Equation \ref{eqn:xi}, and $T_{min}$ and $T_{max}$ are user-defined hyper-parameters that specify the range for the softmax temperature.

Table \ref{tab:CST} provides the top-1 test accuracies for CIFAR-10, 4K CIFAR-10, CIFAR-100, and ImageNet when training with cyclical softmax temperatures.
Our experiments include using a constant $T = 1$ in softmax (column 2),  and ranges of $T = 0.75$ to 1.5 (column 3), $T = 0.5$ to 2 (column 4), and $T = 0.33$ to 3 (column 5).  In addition, we found that the best performance was obtained with a value of $f_c = 1$, which is what we used in our experimental results shown in Table \ref{tab:CST}.
In all four datasets, using a range of temperatures we obtained a modest improvement over the constant $T = 1$ results.  We obtained our best results for a temperature range of $T = 0.5 - 2$, and the performance usually suffered when the range was increased to $T = 0.33 - 3$.  This is an example of how a small amount of range in cyclical training is good but too much can harm the performance.

The implementation of cyclical softmax temperature is straightforward and PyTorch code is available at \url{https://github.com/lnsmith54/CFL} to aid reproducibility.

\section{Data-Based Examples}
\label{sec:data}

Data is essential for training neural networks; the characteristics of the data greatly impact the training.
There is a recent focus in the deep learning community on training data and data augmentation: Andrew Ng launched a campaign for data-centric machine learning and held his first data-centric AI competition\footnote{More information is available at \url{https://https-deeplearning-ai.github.io/data-centric-comp/}}.

The data-based examples of general cyclical training discussed here primarily include varying the use of data augmentation methods and varying which data are used during training.  In addition, we discuss in this Section the proposed technique of varying the amount of unlabeled data included during training for semi-supervised learning.

\subsection{ Data Augmentation}
\label{subsec:DA}

Data augmentation is used widely and includes numerous techniques for improving the generalization ability of deep networks by transforming the training data in ways that do not change the associated label.  
A substantial amount of work has gone into the topic of data augmentation and finding automated ways to find the optimal amount and types of data augmentation for training a network on a specific dataset \cite{shorten2019survey,bayer2021survey,cubuk2020randaugment,zhang2017mixup,devries2017improved,cubuk2018autoaugment}.
However, the previous methods assume that the amount of data augmentation used is constant from the start to the end of training.

Here, we argue that a better technique is to start and end without any data augmentation (or only weak augmentation) and to use an increasing and then decreasing strength of augmentation in the middle epochs.  This follows naturally from the premise of this paper: to ease and encourage learning in the earliest iterations or epochs, then gradually to increase the span of the problem space of the network's learning as the training progresses. This implies starting with no or only very weak augmentation (i.e., image flips) and gradually incorporating stronger augmentation as the training progresses.  In addition, as one approaches the end of the training, it is intuitive to eliminate the strong data augmentation and to fine-tune in the final epochs on the original training data to encourage learning of more prototypical patterns \cite{you2019does}.

One important question that needs to be explored is how to assign data augmentation transformations on a scale from weak to strong.  Intuitively, it is reasonable to measure a transformation by the degree that the transformation modifies the original image's statistics.  Some have assigned image flip and shift transformation as weak augmentation, while strong augmentation includes a number of the image transformations included in the Python Image Library \cite{cubuk2018autoaugment}, mixup \cite{zhang2017mixup}, and cutout \cite{devries2017improved}.  

While it is natural at this point in a research paper to test our hypothesis on cyclical data augmentation (CDA), as a higher level position paper that covers several specific manifestations of the general cyclical training concept, we leave the investigation of a metric for the strength of data augmentation (from the perspective of the trained network) and the testing of CDA for future work.

\subsection{Data}
\label{subsec:data}

Not every training example contributes to a network's learning in the same way.  Training data examples can differ from each other in quality, how representative they are of their class, and uniqueness.  A training sample might have distracting backgrounds, corruptions, or a poor foreground/background ratio.  For these reasons and more, training samples have been divided into categories of easy/medium/hard \cite{zhou2021samples,hendrycks2019benchmarking}. 
This has led to the research question: Which training samples should be learned first and then in what order?  

Kumar, \etal \cite{kumar2010self} propose that the order is determined by how easy the samples are to learn.  
The authors propose a self-paced learning algorithm where training starts with only easy samples and the number of samples increases with each epoch until all the training data is used.  They define easy as ``a set of samples is easy if it admits a good fit in the model space''. The authors demonstrated on a SVM that their learning algorithm outperformed training on all of the training data.

This algorithm is an example of data-based cyclical training, which corresponds to setting $f_c = 1$ in Equation \ref{eqn:xi}.  We argue that this self-paced learning algorithm would outperform training on  all the training data for deep networks based on the same logic in this earlier work \cite{kumar2010self}.  In addition, Kumar, \etal provides evidence that a data-based cyclical training approach likely would be at least comparable to training on a constant training on all the training data.  
%Simultaneous and independent of our work, Kesgin and Amasyali proposed cyclical curriculum learning of training data \cite{kesgin2022cyclical}.

One of the challenges of data-based cyclical training is to automatically measure how easy or hard each training sample is to learn while training the network.  Such a metric should be with respect to the learner at its current competency level.  

This is an open question but we suggest that each sample's loss can be used as one such measure.  Since it is desirable in the early epochs to reduce the contributions to the loss of harder examples with high loss, this implies that a cyclical gradient clipping (CGC) method could improve a network's training.  Specifically, the clipping threshold would be relatively low at the start of training, would increase to a large value in the middle epochs, and gradually would decrease to a low threshold at the end of the training epochs.  A small variation on this method is to set to zero the contribution to the loss of any sample exceeding this cyclical threshold rather than clipping the loss, which would be equivalent to eliminating the hard samples from the training.  Therefore, this variation to cyclical gradient clipping is one way to implement cyclical data training.

Table \ref{tab:CGC} shows results for cyclical gradient clipping that follows the formula:
\begin{equation}
	C = \xi GC_{min} + (1 - \xi) GC_{max}
	\label{eqn:CGC}
\end{equation}
where $C$ is the gradient clipping threshold, $\xi$ follows Equation \ref{eqn:xi}, and $GC_{min}$ and $GC_{max}$ are user-defined hyper-parameters that specify the range of values to use for clipping.  The first column of this Table gives the dataset name, the second gives the accuracy without clipping, and the third is the performance for cyclical gradient clipping.  In this example, we used a clipping mode of value and set $GC_{min} = 4$ and $GC_{max} = 10$. In these experiments $f_c = 2$ was used. These results show a small improvement with cyclical clipping.  We did not test the variation discussed above where the contribution of samples exceeding the clipping threshold is set to zero, which we leave for future work.  

Finally, we note that some researchers suggest that better results can be obtained by training with hard samples first \cite{zhou2021samples}.  These authors demonstrate this for imbalanced datasets, where they define ``hard'' as the same as rare examples.  This is fundamentally different from how Kumar, \etal or we define ``easy/hard''.  In the case of training a highly imbalanced dataset, it is reasonable to include rare examples from the beginning. The way one defines easy or hard is crucial when stating this question.

%One of the distinctions between easy and hard training samples is based on the intra-class and inter-class distances. Easy samples are prototypical of their class and they minimize the inter-class confusion.  Hard samples are near the class decision boundaries and have high inter-class confusion.  

%====================================================================================
\begin{table}[htb]
	\caption{\textbf{Cyclical gradient clipping:} Comparison of the test classification accuracies for CIFAR-10, CIFAR-100, and ImageNet.  The first column gives the dataset name, the second gives the accuracy without clipping, and the third gives the performance for cyclical gradient clipping with a clipping minimum value of 4 (mode is value) and a maximum value of 10 ($f_c = 2$).  These results show an improvement with cyclical gradient clipping.
	 }
	\label{tab:CGC}
	%	\vskip 0.15in
	\begin{center}
		\begin{tabular}{|l|c|c|}
			%					\toprule
			\hline
			Dataset & No clipping & Clip = 4 to 10    \\
			\hline
			%					Model &  &  &  &   &   \\
			%					\midrule
			CIFAR-10 & 97.33$\pm$ 0.07 & 97.48$\pm$ 0.03  \\
			\hline
			CIFAR-100 & 83.82$\pm$ 0.26 & 84.11$\pm$ 0.05  \\
			\hline
			ImageNet & 80.27$\pm$ 0.01& 80.38$\pm$ 0.02 \\
			\hline
			%					\bottomrule
		\end{tabular}
	\end{center}
	\vskip -0.1in
\end{table}
%====================================================================================

\subsection{Semi-supervised Learning}
\label{subsec:SSL}

Semi-supervised learning is a hybrid between supervised and unsupervised learning, which combines the benefits of both \cite{smith2020building}.  As with supervised learning, semi-supervised learning defines a task (i.e., classification) from labeled data, but typically, it requires many fewer labeled samples than supervised learning by leveraging feature learning from unlabeled data to avoid overfitting the limited labeled samples.

Based on the general cyclical concept, this paper proposes a cyclical semi-supervised learning approach  where one starts training the network with only the small labeled dataset and increasingly includes learning with unlabeled samples in the first part of training.  In the second part of the training, the learning with unlabeled samples is decreased gradually so that the training ends with training only on the labeled target data (i.e., fine-tuning on the labeled data, possibly with weak data augmentation). We leave the testing of this technique to future work.

\section{Model-Based Examples}
\label{sec:model}

The principle of general cyclical training can extend to the network's architecture.  Although it is currently unusual for the architecture to change while training, there exists work on growing a network during training \cite{smith2016gradual,chen2015net2net,srivastava2015training}.  However, it is more practical to consider the the general cyclical training principle in the scenarios involving more than one network, such as with transfer learning/pre-training and with teacher-student/knowledge distillation \cite{wang2021knowledge,hinton2015distilling}.

This Section discusses how cyclical training concepts imply procedural changes to training in two important domains:  pre-training a network and knowledge distillation.  We leave the testing of the techniques discussed in this Section to future work.

\subsection{Transfer Learning and Pre-training}
\label{subsec:pretrain}

A common technique used in situations involving a limited number of labeled training data is to pre-train the network's weights on a large, labeled source dataset that has similar characteristics to the target dataset or to use unsupervised pre-training when there’s a large number of unlabeled samples. Pre-training is similar to transfer learning when a network's weights were trained on another dataset and are used to initialize the weights and then the model is fine-tuned on the target dataset.  In both cases, the goal is to maximize the performance on the target data, rather than maximizing the source data performance.

General cyclical training provides guidance on techniques for pre-training.  It is intuitive that the pre-training stage represents the first part of the cycle, where the training should start easy (i.e., HP, data augmentation, and loss), and fully span the problem space as the training proceeds (i.e., to be inclusive of the target data's features).  This is equivalent to using $f_c = 1$ in Equation \ref{eqn:xi} when pre-training. Fine-tuning on the target dataset is equivalent to the second half of the cycle, which implies a large value for $f_c$.  Therefore, it is wise to consider the two training steps (i.e., pre-training and fine-tuning) as part of a single training cycle.

For example, if one is training for the purpose of transfer learning, one should start training in an easy manner (to encourage the weights to move in an optimal direction) and end with hard training. Specifically, one can use the techniques described in this paper with $f_c = 1$, such as cyclical softmax temperature, cyclical data augmentation, and cyclical weight decay.  This should prepare the learned features better than the standard training methodology.  Then one can employ a fine-tuning methodology with the target data, such as using a small learning rate and minimizing the use of strong data augmentation. 

\subsection{Knowledge Distillation}
\label{subsec:KD}

Knowledge distillation was described by Hinton, \etal \cite{hinton2015distilling} using two networks for model compression: the teacher and student networks.  Usually, the student network is smaller than the teacher network and the goal is network compression, in which the smaller student network achieves similar performance as the teacher.

In student-teacher model compression, it is desirable that the student networks should learn as much as possible from the teacher \cite{wang2021knowledge}, with the goal of maximizing the mutual information between the teacher and student.  For this reason, the student loss function often contains not simply a cross-entropy softmax for the labels, but also loss terms for the student to match the teacher's features (the features from the teacher's hidden layers are also called ``dark information'').  This includes deep supervision, in which features at several layers are matched and not just the final hidden layer's features \cite{romero2014fitnets,zhang2019your}.  

Along these lines, it is helpful when training the student to diversify the training characteristics.  This can include using cyclical data augmentation so the student is exposed to strong augmentation during the middle epochs (in addition to weak augmentation for the early and final epochs).  It also can include using cyclical softmax temperature, where the higher temperatures expose the student to what the teacher considers as other close classes in addition to training the student to classify the target class.  
%The use of cyclical methods in training the student adheres to the principle of curriculum learning during student training.

We note that there exists work on teacher-student models in which the teacher chooses what the student should learn \cite{jiang2018mentornet,portelas2020teacher}.  The teacher, in these cases, is guided by curriculum learning in the choice of data or task for the student to learn; these techniques also can be modified to be cyclical.

%Teacher-student methods where the student is more complex than the teacher.   Distillation, where the student is smaller than the teacher.
%[teacher-student Curriculum learning, Mentornet, Teacher algorithms for curriculum learning of Deep RL in continuously parameterized environments]

\section{Discussion}
\label{sec:discussion}

This paper introduces the concept of general cyclical training, which we define as any collection of techniques where training starts and ends with ``easy training'' and the ``hard training'' happens during the middle epochs.
In other words, we can consider a combination of curriculum learning in the early epochs, training on the full expanse of the problem space in the middle epochs, and fine-tuning at the end as general cyclical training. 
In addition, general cyclical training is analogous to curriculum learning in the sense that numerous specific techniques can embody its principles.

This paper specified several novel techniques that follow this principle: cyclical weight decay, cyclical batchsize, cyclical softmax temperature, cyclical data augmentation, cyclical gradient clipping, and cyclical semi-supervised learning.  Each of these techniques is unique, but they are all manifestations of the same concept.  We provided empirical evidence for some of these techniques by showing that they can improve the test performance of trained models, which is also evidence of the validity of the general cyclical training concept.

Although several novel methods are described in this paper, these are only a few of the many potential cyclical techniques that are possible.  Curriculum learning can be considered a component of cyclical training and at the time of this writing, there are well over 3,000 citations to Bengio, \etal \cite{bengio2009curriculum}: how many of these can be converted to a cyclical technique? 

Although cyclical methods introduce new hyper-parameters, one of the benefits of cyclical training methods is a potential overall reduction of the amount of hyper-parameter tuning that is required. In this paper, we have demonstrated that the cliche of ``close'' applies in neural network training (in addition to the game of horseshoes and with hand grenades).  That is, if the optimal training hyper-parameters fall within the cyclical ranges, even if not in the center, the trained network's performance is generally optimal.  When using ranges rather than specific values, a grid search will require fewer tests.
%We hypothesize that it is possibly optimal to allow the learning rate, weight decay, batch size, and momentum a small amount and in concert with Equation \ref{eqn:HPratio}.

The intended purpose of this paper is for the reader to find general cyclical training enlightening, that it illustrates relationships between previously separate methods, and that it encourages the creation of additional novel techniques based on its principles.  We hope these goals were reached.

%\section*{Software and Data}
%
%If a paper is accepted, we strongly encourage the publication of software and data with the
%camera-ready version of the paper whenever appropriate. This can be
%done by including a URL in the camera-ready copy. However, \textbf{do not}
%include URLs that reveal your institution or identity in your
%submission for review. Instead, provide an anonymous URL or upload
%the material as ``Supplementary Material'' into the CMT reviewing
%system. Note that reviewers are not required to look at this material
%when writing their review.
%
%% Acknowledgements should only appear in the accepted version.
\section*{Acknowledgements}
We thank the US Naval Research Laboratory for supporting this research.  
%The views, opinions and/or findings expressed are those of the authors and should not be interpreted as representing the official views or policies of the Department of Defense or the U.S. Government.
%
\bibliographystyle{unsrt}  
\bibliography{GCT}

%%%%%%%%%%%%%%%%%%%%%%%%%%%%%%%%%%%%%%%%%%%%%%%%%%%%%%%%%%%%%%%%%%%%%%%%%%%%%%%
%%%%%%%%%%%%%%%%%%%%%%%%%%%%%%%%%%%%%%%%%%%%%%%%%%%%%%%%%%%%%%%%%%%%%%%%%%%%%%%
% APPENDIX
%%%%%%%%%%%%%%%%%%%%%%%%%%%%%%%%%%%%%%%%%%%%%%%%%%%%%%%%%%%%%%%%%%%%%%%%%%%%%%%
%%%%%%%%%%%%%%%%%%%%%%%%%%%%%%%%%%%%%%%%%%%%%%%%%%%%%%%%%%%%%%%%%%%%%%%%%%%%%%%
\newpage
\appendix
\onecolumn

\section{Software and Implementation}

We used PyTorch Image Models (timm) \cite{rw2019timm} as a framework in our experiments on CIFAR and ImageNet.  This framework provides the models and downloads the data used in our experiments.  
The original code is available at \url{https://github.com/rwightman/pytorch-image-models}.  The file \url{train.py} was modified by inserting additional several new input parameters via calls to \texttt{add\_argument} and adding a few lines of code for cyclical weight decay (CWD), cyclical softmax temperature (CST), and cyclical gradient clipping (CGC).

Specifically, implementing CWD and CGC involves including the following in the training loop:
\begin{lstlisting}
if args.wd_min > 0 or args.clip_min > 0:
    if args.cyclical_factor*epoch < num_epochs:
        eta = 1.0 - args.cyclical_factor *epoch/(num_epochs-1)
    elif args.cyclical_factor == 1.0:
        eta = 0
    else:
        eta = (args.cyclical_factor*epoch/(num_epochs-1) - 1.0) /
              (args.cyclical_factor - 1.0)
if args.wd_min > 0:
	optimizer.param_groups[0]['weight_decay'] = 
		(1 - eta)*args.wd_max + eta*args.wd_min
elif args.clip_min > 0:
	args.clip_grad = (1 - eta)*args.clip_max + eta*args.clip_min
\end{lstlisting}

Similarly, implementing CST can be performed as follows:
\begin{lstlisting}
if args.T_min > 0:
    if args.cyclical_factor*epoch < args.epochs:
	    eta = 1.0 - args.cyclical_factor *epoch/(args.epochs-1)
	elif args.cyclical_factor == 1.0:
 	    eta = 0
	else:
	    eta = (args.cyclical_factor*epoch/(args.epochs-1) - 1.0) /
	          (args.cyclical_factor - 1.0)
	Temperature = (1 - eta)*args.T_max + eta*args.T_min
	output = torch.div(output, Temperature)
\end{lstlisting}
The full revised \texttt{train.py} is available as part of our Supplemental Materials.

%====================================================================================
\begin{table*}[tb]
	\caption{The hyper-parameters used for experiments whose results are presented in the main paper.}
	\label{tab:HP}
	%	\vskip 0.15in
	\begin{center}
		\begin{small}
			\begin{sc}
				\begin{tabular}{|l|c|c|c|c|c|c|}
					%					\toprule
					\hline
					Table & Dataset & Model & Batch size & LR & WD & $f_c$ \\
					\hline
					Table \ref{tab:WD} & CIFAR-10 & TResNet\_m  & 384 & 0.15 & $5 \times 10^{-4} $ & 2, 2, 2  \\
					\hline
					Table  \ref{tab:WD} & CIFAR-100 & TResNet\_m  & 64 & 0.2 & $2 \times 10^{-4} $ & 2, 2, 2  \\
					\hline
					Table  \ref{tab:WD} & ImageNet & TResNet\_m  & 192 & 0.6 & $2 \times 10^{-5} $ & 1,4,4  \\
					\hline
					Table \ref{tab:CST} & CIFAR-10 & TResNet\_m  & 384 & 0.15 & $5 \times 10^{-4} $ & 1, 1, 1  \\
					\hline
					Table  \ref{tab:CST} & CIFAR-100 & TResNet\_m  & 64 & 0.2 & $2 \times 10^{-4} $ & 1, 1, 1  \\
					\hline
					Table  \ref{tab:CST} & ImageNet & TResNet\_m  & 192 & 0.6 & $2 \times 10^{-5} $ & 1, 1, 1  \\
					\hline
					Table \ref{tab:CGC} & CIFAR-10 & TResNet\_m  & 384 & 0.15 & $5 \times 10^{-4} $ & 2  \\
					\hline
					Table  \ref{tab:CGC} & CIFAR-100 & TResNet\_m  & 64 & 0.2 & $2 \times 10^{-4} $ & 2  \\
					\hline
					Table  \ref{tab:CGC} & ImageNet & TResNet\_m  & 192 & 0.6 & $2 \times 10^{-5} $ & 2  \\
					\hline
					%					\bottomrule
				\end{tabular}
			\end{sc}
		\end{small}
	\end{center}
	\vskip -0.1in
\end{table*}
%====================================================================================

\section{Command Lines and Hyper-parameters}

In the spirit of easy replication, it is important to know the values of the hyper-parameters used.  Table \ref{tab:HP} specifies the batch sizes, learning rates, and weight decay values used for the results in Table \ref{tab:WD}, Table \ref{tab:CST} and Table \ref{tab:CGC} in the main body of the paper.

Here, we present the command line for submitting an experiment on cyclical softmax temperature (CST) with Imagenet:
\begin{verbatim}
./distributed_train.sh 4 data/imagenet -b=192 --lr=0.6 --warmup-lr 0.02 
--warmup-epochs 3 --T_min 0.5 --T_max 2 --weight-decay 2e-5 --cooldown-epochs 1 
--model-ema --checkpoint-hist 4 --workers 8 --aa=rand-m9-mstd0.5-inc1 -j=16 --amp 
--model=tresnet_m --epochs=200 --mixup=0.2 --sched='cosine' --reprob=0.4 
--remode=pixel  --cyclical_factor 1 
\end{verbatim}
Default values for hyper-parameters not specified in this command line were used (i.e., see the software in the Supplemental Materials or the original code at \url{https://github.com/rwightman/pytorch-image-models} for default values of the hyper-parameters).  The new hyper-parameters for CST (i.e., T\_min and T\_max) are specified on the command line.  For cyclical weight decay (CWD), the command line was modified by replacing:
\begin{verbatim}
--T_min 0.5 --T_max 2 --weight-decay 2e-5 
--cyclical_factor 1 
\end{verbatim}
with:
\begin{verbatim}
--wd_min 1e-5 --wd_max 8e-5 
--cyclical_factor 2
\end{verbatim}

For CIFAR-10, the following command line was used for CST:
\begin{verbatim}
CUDA_VISIBLE_DEVICES=0 python train.py  data/cifar10 --dataset torch/cifar10
-b 384 --model tresnet_m --checkpoint-hist 4 --sched cosine --epochs 200 --lr 0.5 
--warmup-lr 0.01 --warmup-epochs 3 --cooldown-epochs 1 --weight-decay 5e-4 
--T_min 0.5 --T_max 2 --amp --remode pixel --reprob 0.6 --aug-splits 3 
--aa rand-m9-mstd0.5-inc1 --resplit --split-bn  --dist-bn reduce  --cyclical_factor 1
\end{verbatim}

For CIFAR-100, the following command line was used for CST:
\begin{verbatim}
	CUDA_VISIBLE_DEVICES=0 python train.py  data/cifar100 --dataset torch/cifar100 
	-b 64 --model tresnet_m --checkpoint-hist 4 --sched cosine --epochs 200 --lr 0.2 
	--warmup-lr 0.01 --warmup-epochs 3 --cooldown-epochs 1 --weight-decay 2e-4 
	--T_min 0.5 --T_max 2 --amp --remode pixel --reprob 0.6 --aug-splits 3 
	--aa rand-m9-mstd0.5-inc1 --resplit --split-bn  --dist-bn reduce  --cyclical_factor 1
\end{verbatim}

As with ImageNet, for cyclical weight decay (CWD), each of the above command lines were modified by replacing:
\begin{verbatim}
	--T_min 0.5 --T_max 2 --weight-decay 2e-5 
	--cyclical_factor 1 
\end{verbatim}
with:
\begin{verbatim}
	--wd_min 1e-5 --wd_max 8e-5 
	--cyclical_factor 2
\end{verbatim}
The default values for hyper-parameters are available in the software.

%%%%%%%%%%%%%%%%%%%%%%%%%%%%%%%%%%%%%%%%%%%%%%%%%%%%%%%%%%%%%%%%%%%%%%%%%%%%%%%

\end{document}